\documentclass[sigconf, nonacm]{acmart}
\usepackage{xcolor}
\usepackage[multiple]{footmisc}
\usepackage[dvipsnames]{xcolor}
\usepackage{multirow}
\AtBeginDocument{%
  }
\begin{document}
\title[Leveraging Spatial-Temporal Graph Neural Networks for Multi-Store Sales Forecasting]{Leveraging Spatial-Temporal Graph Neural Networks for Multi-Store Sales Forecasting}
\author{Manish}
\email{23b0354@iitb.ac.in}

\affiliation{%
  \institution{Department of Chemical Engineering, IIT Bombay}
  \city{Mumbai}
  \state{Maharashtra}
  \country{India}
}
\authornote{All the authors contributed equally to this research.}

\author{Arpita Dayama}
\authornotemark[1]
\email{23b0423@iitb.ac.in}
\affiliation{%
    \institution{Department of Chemical Engineering, IIT Bombay}
  \city{Mumbai}
  \state{Maharashtra}
  \country{India}
  }

\iffalse
\author{Aparna Patel}
\affiliation{%
 \institution{Rajiv Gandhi University}
 \city{Doimukh}
 \state{Arunachal Pradesh}
 \country{India}}

\author{Huifen Chan}
\affiliation{%
  \institution{Tsinghua University}
  \city{Haidian Qu}
  \state{Beijing Shi}
  \country{China}}

\author{Charles Palmer}
\affiliation{%
  \institution{Palmer Research Laboratories}
  \city{San Antonio}
  \state{Texas}
  \country{USA}}
\email{cpalmer@prl.com}

\author{John Smith}
\affiliation{%
  \institution{The Th{\o}rv{\"a}ld Group}
  \city{Hekla}
  \country{Iceland}}
\email{jsmith@affiliation.org}

\author{Julius P. Kumquat}
\affiliation{%
  \institution{The Kumquat Consortium}
  \city{New York}
  \country{USA}}
\email{jpkumquat@consortium.net}
\fi

\begin{abstract} 
    This study investigates the effectiveness of Graph Neural Networks (GNNs) for retail sales forecasting across multiple stores and compares their performance with three established baselines: ARIMA, XGBoost, and LSTM. Using weekly data for 45 Walmart stores, the models were evaluated using MAE, RMSE, and MAPE. The GNN achieved the highest overall accuracy, with a MAPE of 2.08 percent for Store 37, representing nearly a 50 percent improvement over ARIMA. XGBoost performed strongly for stores displaying limited correlation with their neighbours, while LSTM exhibited smoothing behaviour that reduced accuracy for volatile stores. The results support the use of GNN architectures for forecasting tasks involving interconnected retail environments.
\end{abstract}

%%

%% Keywords. The author(s) should pick words that accurately describe
%% the work being presented. Separate the keywords with commas.
%% A "teaser" image appears between the author and affiliation
%% information and the body of the document, and typically spans the
%% page.
% \iffalse
% \begin{teaserfigure}
%   \includegraphics[width=\textwidth]{sampleteaser}
%   \caption{Seattle Mariners at Spring Training, 2010.}
%   \Description{Enjoying the baseball game from the third-base
%   seats. Ichiro Suzuki is preparing to bat.}
%   \label{fig:teaser}
% \end{teaserfigure}
% \fi
% \iffalse
% \received{20 February 2007}
% \received[revised]{12 March 2009}
% \received[accepted]{5 June 2009}
% \fi
%%
%% This command processes the author and affiliation and title
%% information and builds the first part of the formatted document.
\maketitle
\section{Introduction}
Accurate demand forecasting is a central requirement in retail operations, influencing inventory control, staffing, supply chain efficiency, and financial planning. Traditional forecasting approaches often model each retail store independently, which prevents the exploitation of shared trends and correlated behaviour across locations. In modern retail networks, stores are affected by common economic environments, weather conditions, and holiday-driven fluctuations. These shared influences suggest that a multi-store forecasting framework should incorporate inter-store relationships.\newline\newline
Graph Neural Networks present a promising architecture that enables relational modelling through message passing between nodes. By representing stores as nodes connected by similarity-based edges, GNNs can capture spatiotemporal dependencies that standard univariate or non-relational models cannot. This study evaluates the advantages of this approach through a controlled comparison with ARIMA, XGBoost, and LSTM models.
\newline\newline
Unlike prior retail forecasting studies, we explicitly evaluate the contribution
of relational information by comparing a spatiotemporal GNN with strong
non-relational baselines under identical feature engineering and data splits.

\section{Related Work}
Classical forecasting methods such as ARIMA models have been applied extensively to univariate retail data~\cite{box2015time}, but their linear nature limits performance in volatile or highly seasonal environments. Tree-based ensemble methods, including XGBoost~\cite{chen2016xgboost}, have shown strong results in many retail forecasting applications due to their ability to model non-linear interactions and incorporate exogenous variables. Deep learning models, particularly LSTM networks~\cite{hochreiter1997long}, have also been used to capture long-term temporal patterns in sales series. Recent research has highlighted the potential of graph-based models for traffic forecasting~\cite{yu2017spatio}, energy demand prediction, and environmental monitoring~\cite{wu2019graph}. However, applications of GNNs in retail forecasting remain limited. This study builds on this emerging direction by providing an empirical comparison across multiple architectures.

\section{Dataset and Feature Engineering}
\label{sec:dataset}

\subsection{Dataset Description}
The study utilizes the Walmart Weekly Sales dataset originally published on Kaggle by Haji~\cite{haji2016walmart}. The dataset contains weekly observations from 45 Walmart stores spanning February 2010 to October 2012. Each observation includes a store identifier, department identifier, date, weekly sales value, a holiday-week indicator, and several regional macroeconomic attributes. Because our forecasting problem is defined at the store level, the departmental records were aggregated to produce a single weekly sales series for each store.

The raw variables available in the dataset include:
\begin{itemize}
    \item \textbf{Store}: unique store identifier,
    \item \textbf{Date}: week-ending date,
    \item \textbf{Weekly\_Sales}: department-level weekly sales in USD,
    \item \textbf{IsHoliday}: binary indicator for holiday weeks,
    \item \textbf{Temperature}: average weekly temperature (in Fahrenheit),
    \item \textbf{Fuel\_Price}: regional weekly fuel price,
    \item \textbf{CPI}: consumer price index,
    \item \textbf{Unemployment}: regional unemployment rate.
\end{itemize}

\subsection{Temporal and Calendar Features}
To incorporate time-dependent structure, the \texttt{Date} column was parsed and sorted per store. Several temporal descriptors were derived, including year, month, week-of-year, and day-of-week. To avoid discontinuities inherent in cyclical variables, sine and cosine encodings were applied to both week-of-year and month.

Holiday information was enriched by explicitly identifying Christmas (December 25) and Thanksgiving (computed as the fourth Thursday of November for each year). Two new indicators were introduced: \texttt{is\_major\_holiday} for Christmas and Thanksgiving weeks, and \texttt{is\_minor\_holiday} for all other holidays designated by the dataset’s \texttt{IsHoliday} field.

\subsection{Lag Features and Rolling Statistics}
Autocorrelation was captured by generating lagged sales values over the following horizons: 1, 2, 3, 7, 14, 28, and 52 weeks. All lag features were computed separately for each store to prevent cross-store leakage.

Longer-term temporal patterns were modeled using trailing rolling window statistics. Rolling means were computed over 3-, 4-, 8-, 12-, and 52-week windows; rolling standard deviations were computed over 3-, 4-, 8-, and 52-week windows. These statistics were shifted by one period to ensure that only past observations informed the engineered features.

Exponentially weighted moving averages (EWMA) with spans of 3, 7, and 14 weeks were also added, again using only past data via a one-step shift.

\subsection{Sales Transformations}
A smooth target transformation,
\[
\text{sales\_log1p} = \log(1 + \text{Weekly\_Sales}),
\]
was applied to mitigate the effect of extreme spikes during major holiday periods and to stabilize neural network training.

\subsection{Data Cleaning and Store-Level Imputation}
To avoid inter-store contamination, all cleaning and imputation operations were performed within each store independently. Numeric fields were first coerced to NaN, then filled using forward-fill and backward-fill operations per store. Remaining missing values were replaced with zero. The final dataset was re-sorted by \texttt{Store} and \texttt{Date}.

\subsection{Final Feature Set}
The resulting feature matrix includes:
\begin{itemize}
    \item raw economic and temporal variables,
    \item cyclical time encodings,
    \item lagged features across multiple horizons,
    \item rolling means and standard deviations,
    \item EWMA features,
    \item enhanced holiday indicators,
    \item log-transformed sales.
\end{itemize}

\begin{table}[t]
\centering
\caption{Summary statistics of selected raw and engineered features.}
\label{tab:feature_summary}
\begin{tabular}{lrr}
\toprule
Feature & Mean & Std \\
\midrule
Weekly\_Sales (USD) & 1046964 & 564366 \\
sales\_log1p & 13.7 & 0.58 \\
Temperature & 60.6 & 18.4 \\
CPI & 171.57 & 39.35 \\
\bottomrule
\end{tabular}
\end{table}

\begin{figure}[h!]
\centering
\includegraphics[width=\linewidth]{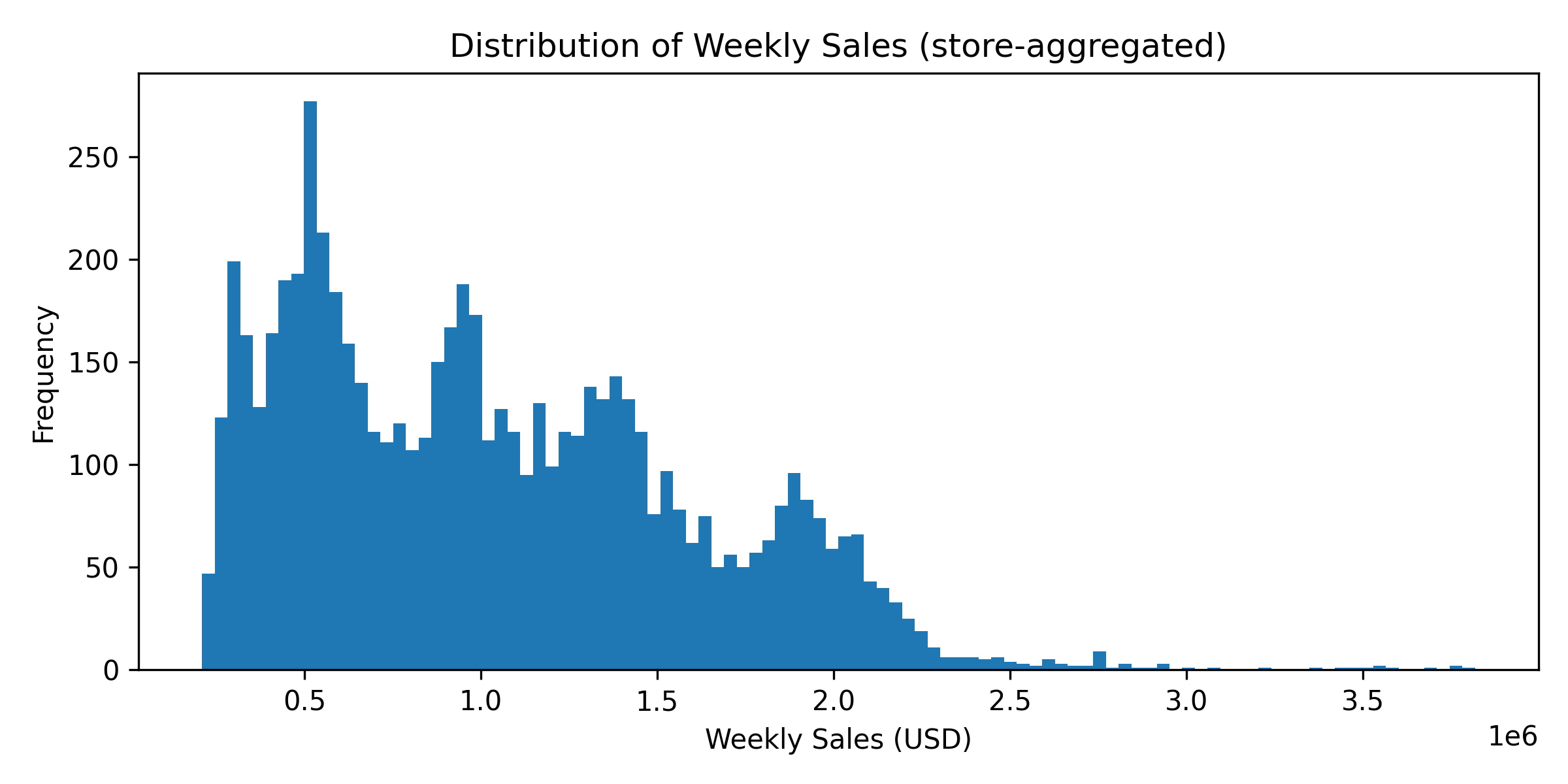}
\caption{Distribution of weekly store-level sales after aggregation.}
\label{fig:sales_dist}
\end{figure}

\begin{figure}[h!]
\centering
\includegraphics[width=\linewidth]{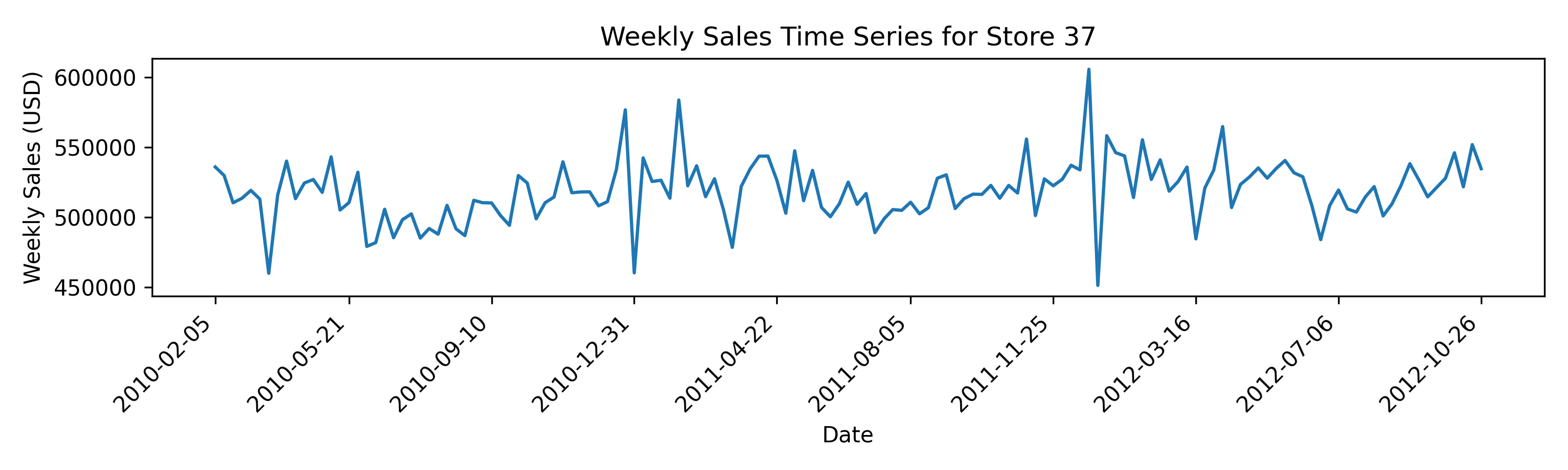}
\caption{Weekly sales time series for a representative store}
\label{fig:sales_time_series}
\end{figure}

\section{Methodology}
\subsection{Traditional and Machine Learning Baselines}

To ensure a fair comparison, all baseline models (ARIMA, LSTM, and XGBoost) were trained using the exact same engineered feature set extracted from features.csv. Only the modeling procedures differed.

\subsubsection{ARIMA (1,0,1) with Exogenous Inputs}

A separate ARIMA model was fit for each store to model raw Weekly Sales using an ARIMA(p,d,q) structure. \newline After preliminary diagnostics, a Fixed ARIMA(1,0,1) specification was used to ensure comparability across stores, which showed no consistent advantage from store-specific orders. All numeric engineered features were provided as exogenous regressors. Models were trained on the first 80 percent of observations and tested on the remaining 20 percent.

\subsubsection{LSTM with Log Transformation}

To stabilize variance, the LSTM model applied a log1p transformation to the target. For stores containing negative values, a per-store positive shift was added before log transformation and removed during inverse conversion.

Input sequences of length 10 were constructed per store and split chronologically into 70 percent training, 15 percent validation, and 15 percent testing. A two-layer LSTM architecture with dropout and dense projections was trained using Huber loss and Adam optimizer. Predictions were inverse-transformed back to dollar values.

\subsubsection{XGBoost Regression}

XGBoost was trained per store using the same engineered features as input and predicting raw weekly sales directly. The model used 1000 estimators, a learning rate of 0.05, and depth 6 with subsampling. Chronological 80/20 train-test splits were applied. No target transformation was used.

\subsection{Spatiotemporal Graph Neural Network (STGNN)}

The STGNN models all 45 Walmart stores jointly and learns both temporal dependencies
within each store and spatial dependencies across stores. Unlike traditional models that
treat each store independently, the STGNN operates on the full multi-store tensor.

\subsubsection{Data Restructuring and Stationary Target}

Weekly sales were arranged as a matrix
\[
Y_{\text{raw}} \in \mathbb{R}^{T \times S},
\]
with \(T\) weeks and \(S = 45\) stores. Sales were transformed into log space:
\[
Y_{\log} = \log(1 + Y_{\text{raw}}).
\]

To obtain a stationary target, the model predicts the log-difference:
\[
Y_{\text{diff}}(t,s) = Y_{\log}(t,s) - Y_{\log}(t-1,s).
\]

The previous log-level
\[
Y_{\text{base}}(t-1)
\]
was stored to reconstruct predictions after inference.

Exogenous features were reshaped to
\[
X \in \mathbb{R}^{T \times S \times F},
\]
standardized using a global scaler, and shifted by one timestep to align with
\(Y_{\text{diff}}\).

\subsubsection{Windowing and Train--Test Splitting}

A rolling window of length 12 was used to form input sequences. The first 80\% of
windows were used for training, and the remaining 20\% for testing. Before entering
the neural network, tensors were permuted to match the PyTorch convention:
\[
(\text{Batch}, \text{Features}, \text{Stores}, \text{Time}).
\]

\subsubsection{Model Architecture}

The STGNN consists of three primary components:

\paragraph{a) Learnable Graph Construction}

A GraphLearner module builds an adaptive adjacency matrix from store embeddings:
\[
A = \text{softmax}\!\left(\text{ReLU}(E_1 E_2^{\top})\right),
\]
allowing the model to infer store-to-store dependencies directly from data.

\paragraph{b) Temporal Feature Extraction}

A dilated Temporal Convolutional Network (TCN)~\cite{bai2018empirical} processes each store’s temporal slice
using stacked \(1 \times 3\) convolutions with dilation. This enables multi-scale temporal
receptive fields while maintaining computational efficiency.

\paragraph{c) Graph Convolution~\cite{kipf2016semi} for Spatial Aggregation}

Spatial aggregation is performed with a learned graph convolution:
\[
Z = A X W,
\]
where \(W\) is a trainable weight matrix. The output is combined with the initial
projection via a residual connection to preserve local store-level temporal patterns.

\paragraph{d) Output Projection}

A final \(1 \times 1\) convolution produces the next-step log-difference for each store.

\subsubsection{Reconstruction to Dollar Sales}

The predicted log-difference is added back to the stored base value:
\[
\hat{Y}_{\log}(t) = Y_{\text{base}}(t-1) + \hat{Y}_{\text{diff}}(t).
\]

The final forecast in dollars is obtained via:
\[
\hat{Y}(t) = \exp(\hat{Y}_{\log}(t)) - 1.
\]

\subsubsection{Training Procedure}

The model was trained for 100 epochs using AdamW~\cite{loshchilov2017decoupled} (learning rate 0.001, weight decay
\(10^{-4}\)) and Smooth L1 loss. A ReduceLROnPlateau scheduler adjusted the learning rate
based on validation loss. Training and validation MAE curves were monitored to verify 
convergence and avoid overfitting.

\subsection{Evaluation Metrics}

To comprehensively assess the model's performance beyond standard error averaging, we utilised four specific metrics: Normalised Total Absolute Error (NTAE), Win Rate, P90 MAPE, and Variance of MAPE.

Let $y_{t,s}$ represent the actual sales value and $\hat{y}_{t,s}$ represent the predicted sales value for store $s$ at time step $t$. Let $S$ be the total number of stores and $T$ be the total number of time steps in the test set.

\subsubsection{Normalized Total Absolute Error (NTAE)}
Calculated globally across all stores and time steps (equivalent to summing all absolute errors and dividing by total actual volume):

\begin{equation}
    \text{NTAE} = \frac{\sum_{s=1}^{S} \sum_{t=1}^{T} |y_{t,s} - \hat{y}_{t,s}|}{\sum_{s=1}^{S} \sum_{t=1}^{T} y_{t,s}} \times 100
\end{equation}

\subsubsection{Store-Level Metrics}
We compute the Mean Absolute Error (MAE) and Root Mean Squared Error (RMSE) individually for each store $s$:

\begin{equation}
    \text{MAE}_s = \frac{1}{T} \sum_{t=1}^{T} |y_{t,s} - \hat{y}_{t,s}|
\end{equation}

\begin{equation}
    \text{RMSE}_s = \sqrt{\frac{1}{T} \sum_{t=1}^{T} (y_{t,s} - \hat{y}_{t,s})^2}
\end{equation}

\subsubsection{Distributional Metrics (P90 and Variance)}
To assess robustness, we first calculate the Mean Absolute Percentage Error (MAPE) for each store $s$:

\begin{equation}
    \text{MAPE}_s = \frac{1}{T} \sum_{t=1}^{T} \left| \frac{y_{t,s} - \hat{y}_{t,s}}{y_{t,s}} \right| \times 100
\end{equation}

Let $\mathcal{M} = \{ \text{MAPE}_1, \text{MAPE}_2, \dots, \text{MAPE}_S \}$ be the set of MAPE values for all stores. We report:

\begin{itemize}
    \item \textbf{P90 MAPE:} The $90^{th}$ percentile of the set $\mathcal{M}$, representing the error threshold for the worst-performing 10\% of stores.
    \item \textbf{Variance of MAPE:} Measures the consistency of model performance across stores:
    \begin{equation}
        \text{Var}(\mathcal{M}) = \frac{1}{S} \sum_{s=1}^{S} (\text{MAPE}_s - \mu_{\mathcal{M}})^2
    \end{equation}
    where $\mu_{\mathcal{M}}$ is the mean of all store MAPEs.
\end{itemize}
\subsubsection{Win Rate (WR)}
The Win Rate measures the frequency with which the proposed model outperforms a persistence baseline (where the prediction $\hat{y}^{\text{base}}_{t,s}$ equals the last observed value). It is calculated as the percentage of individual test instances where the model's absolute error is lower than the baseline's:

\begin{equation}
    \text{WR} = \frac{1}{S \times T} \sum_{s=1}^{S} \sum_{t=1}^{T} \mathbb{I}\left( |y_{t,s} - \hat{y}_{t,s}| < |y_{t,s} - \hat{y}^{\text{base}}_{t,s}| \right) \times 100
\end{equation}

where $\mathbb{I}(\cdot)$ is the indicator function, which returns 1 if the condition is true and 0 otherwise.

% LSTM and STGNN internally used log representations, but all final metrics were computed on real weekly sales.

\section{Experimental Results}
\label{sec:results}

This section reports the empirical results for ARIMA, XGBoost, LSTM, and Graph Neural Network (GNN) models. Performance is measured using MAE, RMSE, and MAPE computed at the store level. Results are reported for representative stores and summarised across models.

\subsection{ARIMA Performance}
ARIMA served as the statistical baseline and displayed relatively high error rates in many stores.

\begin{itemize}
  \item \textbf{Store 4}: MAPE = 4.43\%, MAE = \$95,129, RMSE = \$120,315.
  \item \textbf{Store 10}: MAPE = 7.15\%, MAE = \$127,708, RMSE = \$150,583.
  \item \textbf{Store 36}: MAPE = 3.01\%, MAE = \$9,305, RMSE = \$11,536.
\end{itemize}

Overall, ARIMA underperformed relative to machine learning and graph-based models, largely because it cannot leverage exogenous covariates or cross-store signals.

\subsection{XGBoost Performance}
XGBoost demonstrated robust performance across many stores, often surpassing ARIMA and LSTM.

\begin{itemize}
  \item \textbf{Store 43}: MAPE = 2.22\%, MAE = \$13,731, RMSE = \$16,734.
  \item \textbf{Store 37}: MAPE = 2.48\%, MAE = \$13,024, RMSE = \$16,507.
  \item \textbf{Store 10}: MAPE = 3.3\% (XGBoost outperformed GNN for this store).
\end{itemize}

XGBoost is particularly effective for stores that behave relatively independently of neighbours, benefiting from engineered lag and rolling features.

\subsection{LSTM Performance}
LSTM forecasts tended to be smoothed, which harmed performance on volatile stores.

\begin{itemize}
  \item \textbf{Store 45}: MAPE = 4.91\%.
  \item \textbf{Store 11}: MAPE = 6.55\%.
  \item \textbf{Store 27}: Forecasts were overly smooth while the true series exhibited high-frequency fluctuations.
\end{itemize}

The limited temporal depth per store (approximately 140 weekly observations) likely constrained the LSTM's advantages.

\subsection{GNN Performance}
The GNN attained the best overall results by exploiting inter-store relationships through message passing.

\begin{itemize}
  \item \textbf{Store 37}: MAPE = 2.08\%, MAE = \$10,923, RMSE = \$13,727 (best overall).
  \item \textbf{Store 30}: MAPE = 2.35\%, MAE = \$10,152, RMSE = \$13,397.
  \item \textbf{Store 5}: MAPE = 3.75\%, MAE = \$12,082, RMSE = \$15,158.
\end{itemize}

GNN predictions show low variance between MAE and RMSE, indicating few large-error outliers and stable forecasts in correlated clusters.

\subsection{Learned Spatial Dependencies and Centrality Analysis}

\begin{figure}[h!]
    \centering
    \includegraphics[width=\linewidth]{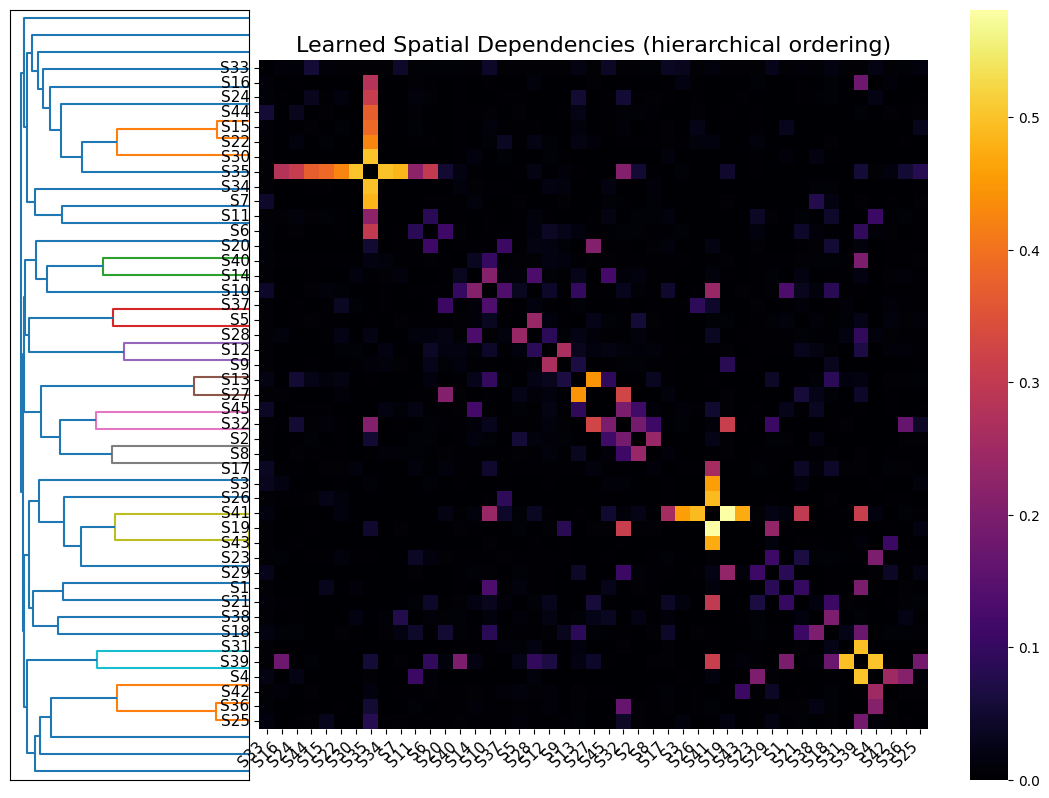} % 
    \caption{Reordered learned adjacency matrix showing functional store clusters and high-influence nodes identified by the STGNN}
    \label{fig:heatmap}
\end{figure}

To interpret the spatial structure learned by the STGNN, we reorder the
adaptive adjacency matrix using hierarchical clustering. The reordered heatmap
reveals clear block-diagonal communities of stores with similar sales dynamics,
as well as several high-intensity cross-patterns corresponding to ``influencer''
stores. This structure emerges without geographic metadata, indicating that the
model learns latent functional relationships directly from the data.

We also compute a simple centrality score by averaging each store's outgoing
attention weights. The top-ranked stores by mean centrality are S39, S29, S18,
S19, and S28, while the least central are S38, S27, S4, S21, and S35.

Notably, some highly central stores (e.g., S29, S28) exhibit poor forecasting
accuracy under both STGNN and XGBoost. This is expected: centrality reflects
\emph{influence on other stores}, not local predictability. These stores act as
strong information sources in the graph, yet their own time series are noisy and
weakly seasonal, making them intrinsically difficult to forecast. Conversely,
isolated stores show limited relational benefit but may remain easier or harder
to predict depending on their temporal characteristics.

Overall, the analysis shows that the STGNN captures meaningful inter-store
relationships and benefits stores embedded in stable relational clusters, while
high-error nodes reflect inherent volatility rather than model limitations.

\subsection{Summary Table}

\begin{table}[h!]
  \centering
  \caption{Representative store-level performance metrics (MAPE in percent; MAE and RMSE in USD).}
  \label{tab:model_comparison}
  \small
  \begin{tabular}{r|ccc|c|ccc}
    \toprule
    \multirow{2}{*}{Store} & \multicolumn{3}{c|}{ARIMA} & XGBoost & \multicolumn{3}{c}{GNN} \\
    & MAPE & MAE & RMSE & MAPE & MAPE & MAE & RMSE \\
    \midrule
    4  & 4.43 & 95,129 & 120,315 & 4.32  & 4.40 & 93,711     & 107,445 \\
    10 & 7.15 & 127,708    & 150,583     & 3.30 & 3.86 & 67,696     & 84,677 \\
    36 & 3.01 & 9,305  & 11,536 & 7.24   & 4.83   & 13,990     & 16,700 \\
    43 & 16.09   & 99,497     & 114,690     & 2.22 & 3.97   & 24,541     & 31,420 \\
    37 & 3.98 & 20,952 & 25,521 & 2.48 & 2.08 & 10,923 & 13,727 \\
    30 & 10.11   & 44,078     & 45,774     & 3.21 & 2.35 & 10,152 & 13,397 \\
    5  & 8.42   & 27,320     & 37,480     & 15.67   & 3.75 & 12,082 & 15,158 \\
    \bottomrule
  \end{tabular}
\end{table}

NTAE, WSR, P90 MAPE, and MAPE variance are computed across all 45 stores.
Lower is better for NTAE, P90, and variance.

We report representative stores for brevity, full store-level metrics are provided in supplementary tables.

\begin{figure}[h!]
\centering
\includegraphics[width=\linewidth]{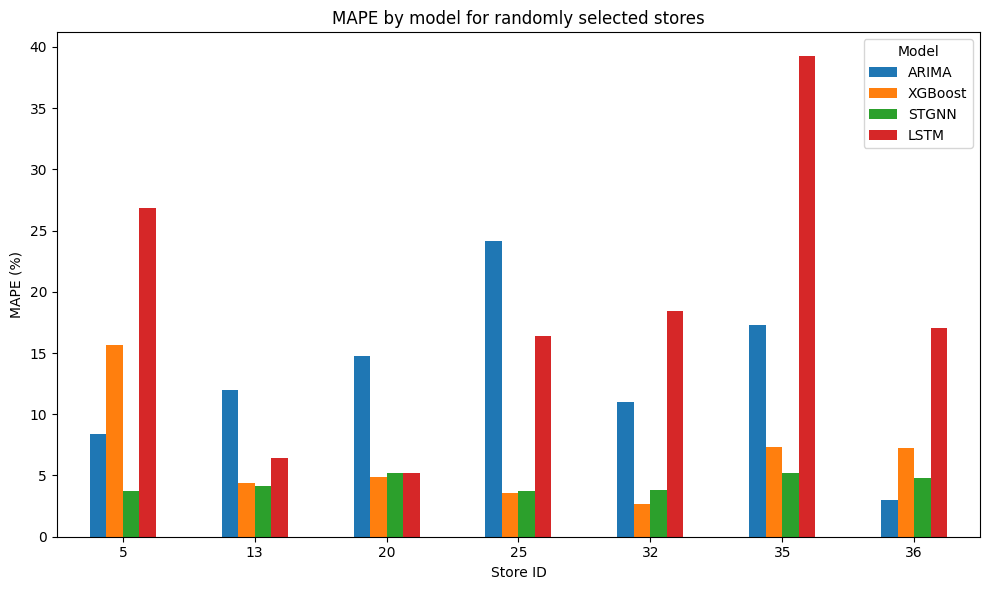}
\caption{MAPE by model for selected stores}
\label{fig:mape_by_model}
\end{figure}

\begin{figure}[h!]
\centering
\includegraphics[width=\linewidth]{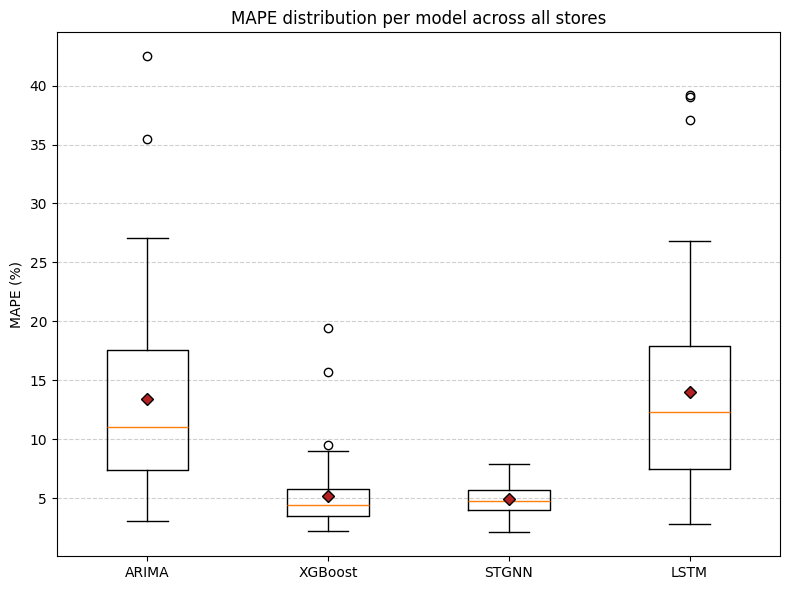}
\caption{MAPE distribution per model across all stores}
\label{fig:mape_box_plot}
\end{figure}

\begin{figure}[h!]
\centering
\includegraphics[width=\linewidth]{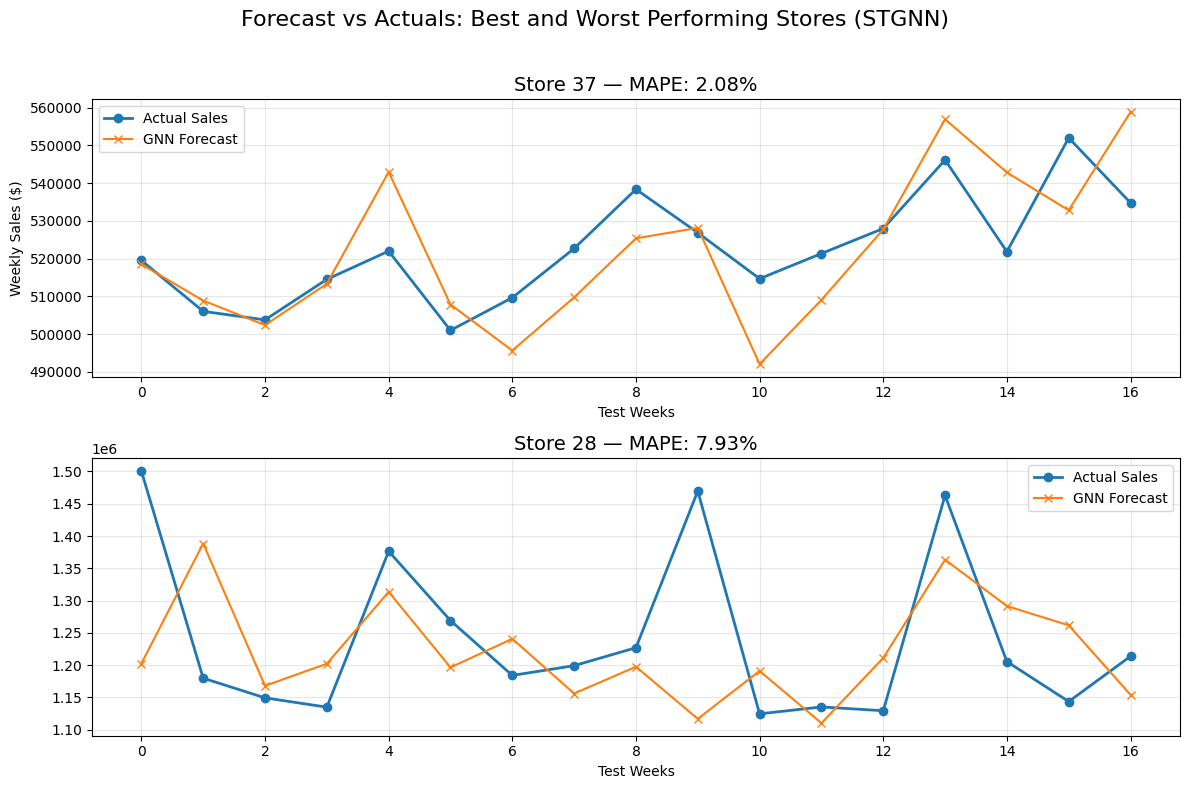}
\caption{Forecast vs actuals for best and worst STGNN stores, illustrating model robustness and failure modes}
\label{fig:xgboost}
\end{figure}

\begin{table}[t]
\centering
\caption{Aggregate forecasting performance across all models}
\label{tab:agg_metrics}
\small
\setlength{\tabcolsep}{3.2pt}
\begin{tabular}{lcccc}
\toprule
\textbf{Model} & 
\textbf{NTAE (\%)} & 
\textbf{Win Rate (\%)} & 
\textbf{P90 MAPE (\%)} & 
\textbf{Var. of MAPE} \\
\midrule
STGNN   & 4.93  & 49.02    & 6.98   & 1.75 \\
XGBoost & 4.99  & 48.17    & 7.30 & 9.79 \\
LSTM    & 12.95 & 18.11    & 21.75  & 72.38 \\
ARIMA   & 13.36 & 22.06    & 24.09  & 70.85 \\
\bottomrule
\end{tabular}
\end{table}

\section{Conclusion}
This work evaluates ARIMA, LSTM, XGBoost, and a Spatiotemporal Graph Neural Network (STGNN) for multi-store sales forecasting and demonstrates that STGNN delivers the most accurate and stable performance across stores. By explicitly modeling inter-store dependencies through a learned graph structure, STGNN outperforms traditional and tree-based baselines, particularly for stores embedded in dense relational clusters.

While XGBoost remains competitive for isolated stores with limited cross-store correlation, its per-store performance variance is noticeably higher than that of STGNN. In contrast, the STGNN achieves lower NTAE, superior P90 error, and the smallest variance in per-store MAPE, indicating greater overall robustness. Future extensions may incorporate dynamic graph structures, attention-based GNNs, and external economic signals.

Future work may extend this research by incorporating dynamic graphs that evolve over time, exploring attention-based GNN architectures to improve interpretability, and integrating external signals such as macroeconomic indicators, promotional calendars, or competitor activity. Additionally, evaluating these models across diverse retail datasets may further validate their generalizability.

In summary, 
Structured tabular forecasting is dominated by tree-based models unless relational structure is present.
GNNs offer robustness and tail-performance benefits that trees lack. Overall, the results demonstrate that relational structure materially improves forecast quality in multi-store environments, establishing STGNNs as a robust and scalable foundation for retail forecasting.

\section{Limitations} 
The dataset has limited temporal depth (140 weeks per store), which restricts the performance of deep temporal models. Several stores exhibit inherently noisy or weakly seasonal behavior, limiting forecastability regardless of model choice. Finally, the learned adjacency matrix is static; future extensions may incorporate dynamic or context-dependent graphs.

\bibliographystyle{ACM-Reference-Format}
\bibliography{ref}

\end{document}